\documentclass[letterpaper, 10 pt, conference]{ieeeconf}  

\IEEEoverridecommandlockouts                              

\usepackage{graphicx} 
\usepackage{algorithm}
\usepackage{algorithmicx}
\usepackage{amsmath}
\usepackage{amsfonts}
\usepackage{algpseudocode}
\usepackage{subfig}
\usepackage{bm}
\newcommand\mycom[2]{\genfrac{}{}{0pt}{}{#1}{#2}}

\title{\LARGE \bf
Collective Decision Making in Communication-Constrained Environments
}

\author{Thomas G. Kelly, Mohammad Divband Soorati, Klaus-Peter Zauner, \\
    Sarvapali D. Ramchurn, Danesh Tarapore
    \thanks{School of Electronics and Computer Science, University of Southampton, Southampton, SO17 1BJ, UK {\tt\small \{tgk2g14, mds1u19, kpz, sdr1, dst1m17\}@soton.ac.uk}}}

\begin{document}

\maketitle
\thispagestyle{empty}
\pagestyle{empty}


\begin{abstract}
One of the main tasks for autonomous robot swarms is to collectively decide on the best available option. Achieving that requires a high quality communication between the agents that may not be always available in a real world environment. In this paper we introduce the communication-constrained collective decision-making problem where some areas of the environment limit the agents’ ability to communicate, either by reducing success rate or blocking the communication channels. We propose a decentralised algorithm for mapping environmental features for robot swarms as well as improving collective decision making in communication-limited environments without prior knowledge of the communication landscape. Our results show that making a collective aware of the communication environment can improve the speed of convergence in the presence of communication limitations, at least 3 times faster, without sacrificing accuracy.
\end{abstract}

\section{Introduction}

Coping with failure is a major challenge for designing adaptive robotic systems. Swarm robotics aims to solve the issue of individual robot failures by using redundancy in the system. A large group of simple robots locally interact with each other and the environment yielding an emerging behaviour. While being characterised as adaptable, scalable and robust, robot swarms face challenges in the design of decentralised collective decision making processes. A collective decision-making strategy has to define a process that solves the \textit{best-of-n} problem where the swarm is tasked to identify the best choice from a set of given options~\cite{valentini2017best}. 

The collective decision making process is a vital element of a swarm system being able to function autonomously \cite{hamann2018swarm}. When the swarm is located in an environment which imposes limitations on the swarm's ability to communicate with one another, which may be in the form of obstructions in the environment (e.g. foliage, electric cables), this can hinder the decision making power of the swarm and may lead to scenarios where a decision takes much longer to be made. In scenarios that are time critical, it is important that decisions are made quickly and accurately by the swarm in order to improve the system's situational awareness \cite{roundtree2019transparency}. For example, consider a swarm that has been tasked with assessing the best route of a finite number of options, to take through a densely packed forest, following a disaster. It is unknown how the existing routes have been affected by the disaster and may pose a risk to humans entering the area.

The collective perception problem was introduced by Valentini et al. \cite{valentini2016collective} where robots are tasked with feature detection. In the problem they posed, the agents are able to observe features in their environment, these being black and white tiles, and must estimate whether there are more white or black tiles present in the environment. Agents maintain an estimate and a belief over which colour is most dominant and are able to share this information with neighbouring agents. Over time, the agents collate their estimates to form a consentaneous belief about which feature is more dominant in the environment. The trade-off between speed and accuracy is examined for different collective decision making strategies, each of which had been previously proposed \cite{valentini2014self, valentini2016100kilobots}. These approaches focused on making use of positive feedback mechanisms to improve the speed at which the swarm converges to a consensus. This positive feedback mechanism was achieved by modulating the time agents spent attempting to share their beliefs and estimates. The collective perception problem was also examined by Ebert et al. where a multi-feature decision making approach was proposed \cite{ebert2018multi}. Here, the authors also constrained the agents to local communications, preventing them from sharing their estimates and beliefs globally across the swarm. 
Against this background, our approach considers a scenario where agents' communication is constrained by the environment. As such, agents must adapt their behaviour based on their current communication conditions. Collective perception is mainly considered in environments with homogeneous communication quality distribution \cite{valentini2014self, valentini2016100kilobots, valentini2016collective}. It has been shown that collective behaviour is possible even without any communication but the emerging behaviour is limited to simple spatial configurations (e.g., aggregation)~\cite{kernbach2013adaptive, schmickl2011beeclust}. Limited range and spatially targeted communications have also been proposed that are design decisions to improve the collective behaviour rather than an externally-imposed constraint~\cite{mathews2010establishing, mathews2012spatially, talamali2021less}. The resilience of different control approaches to communication noise and temporal constraints are considered~\cite{coquet2021control,winfield2006safety} but, to the best of our knowledge, literature lacks a collective decision-making strategy that adapts to environments with inhomogeneous communication quality distribution.

Our study advances the state of the art in the following ways: First, we propose a new benchmark scenario to study the collective perception of a robot swarm in a communication-constrained environment. In our scenario inter-robot communication may be completely denied in a priori unknown areas for the duration of the experiment. Second, we develop algorithms for the swarm  to adapt to these communication restrictions, to achieve a high speed to consensus on decisions of features of interest in the environment.
Third, our results show that our algorithms significantly outperforms the state-of-the-art collective decision making algorithms in communication constrained setups. We next formally define the problem and go on to propose our solution and experiments.

\section{Problem Definition}

\subsection{Collective Perception Problem}

We define a collective perception scenario with 2 features ($f = 2$). Agents are placed randomly within the environment of size $n \times n$ units. Every point in the environment is assigned a value, representing the feature at that location. The task of the agents is to decide which feature is dominant in the environment. The ratio of cells which correspond to a feature of 1 in the environment is given by $r_{f}$. For our experiments, the feature of 1 is always the dominant feature and, as such, the swarm's task is to reach a consensus for 1.


\subsection{Communication-Constrained Environment}

The environment is also characterised by a communication quality, $q_c$, defined for each point, $q_c \in [0,1]$. This quality relates to the probability that a message broadcast by an agent is successfully received by neighbouring agents in communication range. Communication quality varies across the environment. Moreover, the agents are denied communication when in some regions of the environment ($q_c=0$). The ratio of communication denied cells in the environment is given by $r_c$. Agents are able to perceive the value of the feature at their current location in the environment but must estimate the communication link quality themselves.

\begin{figure}[!h]
\centering
\minipage{0.15\textwidth}
\subfloat[]{\includegraphics[trim={3cm 0cm 3cm 1cm},clip,width=\linewidth]{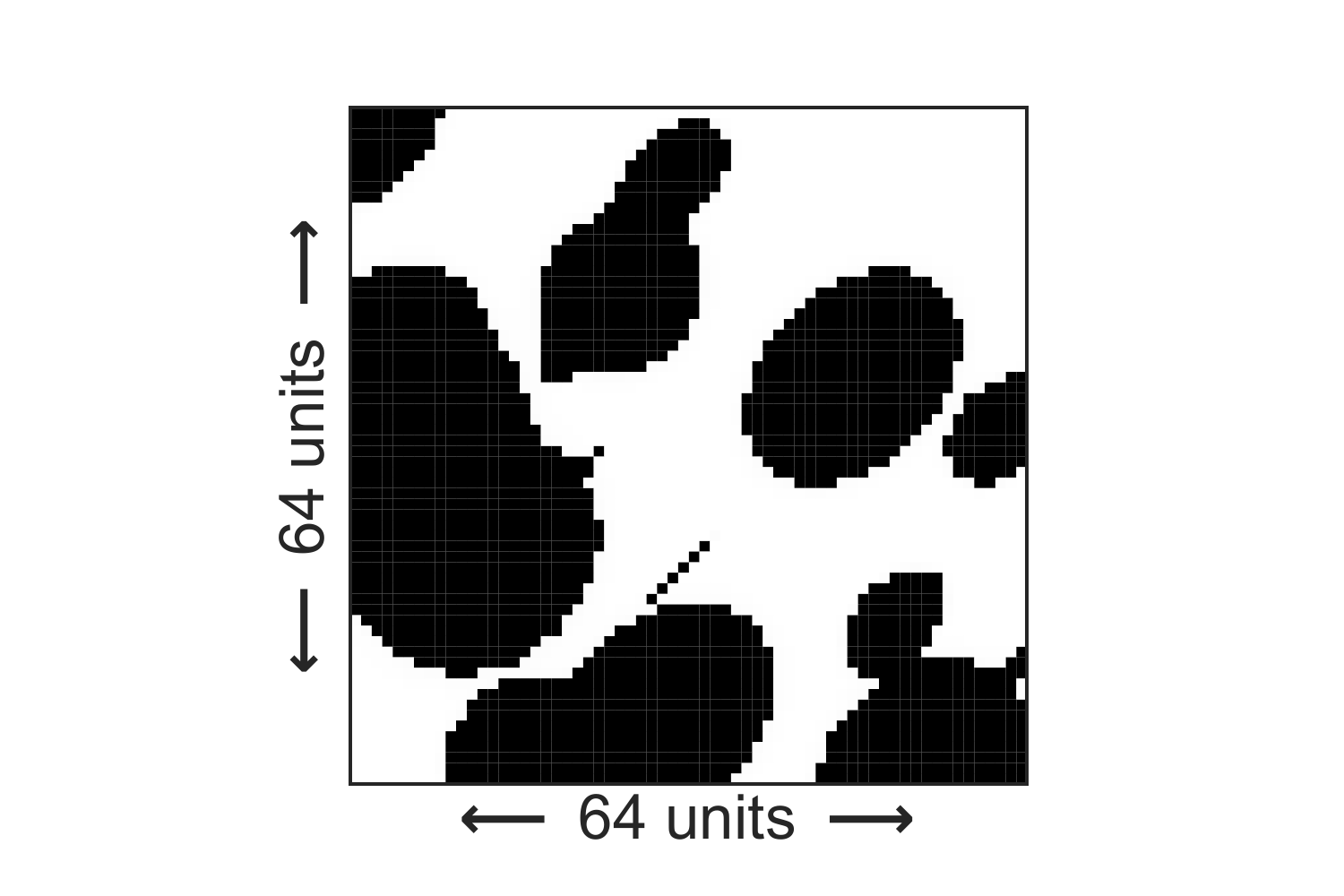}}
\endminipage\hspace{2em}
\minipage{0.20\textwidth}
\subfloat[]{\includegraphics[trim={2cm 0cm 1cm 1cm},clip,width=\linewidth]{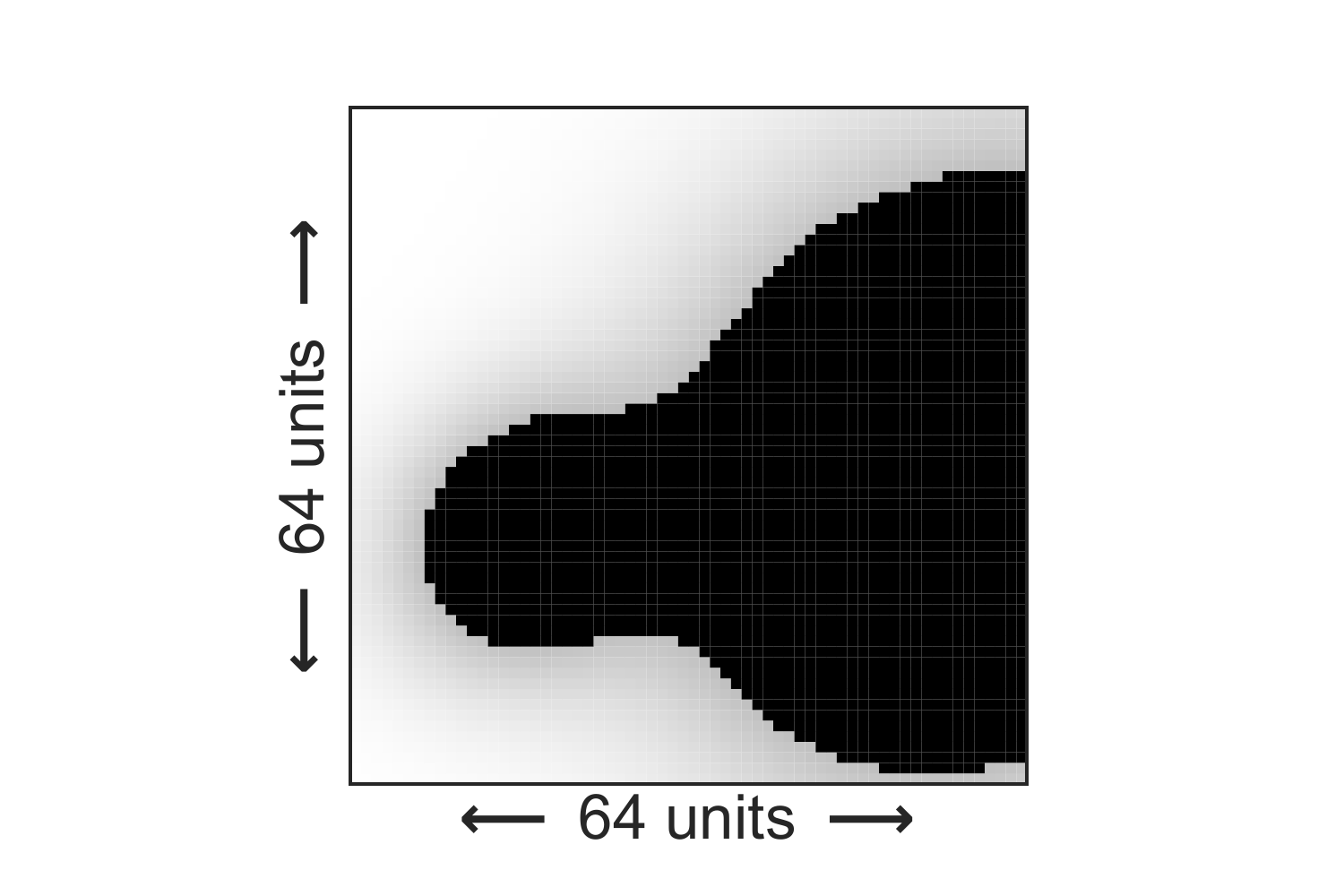}}
\endminipage
\caption{Examples of a distributed communication denied (a), and a continuous communication denied (b) environment. In both environments, agents are denied communication in half the environment ($r_c=0.5$), indicated in black. Areas shaded lighter allow communication with a higher probability.}
\label{fig:commdeniedexamples}
\end{figure}

\section{Method}

Our communication-aware approach consists of three elements. These being the collective decision making strategy on the dominant feature in the environment, augmented with communication-aware path planning, and coordination algorithms for the swarm. In order to evaluate our communication-aware algorithms, we consider two types of communication environments:
1) Distributed: With many, small communication denied areas which are spread throughout the environment, for example, see Figure \ref{fig:commdeniedexamples}a. 2) Continuous: With one large communication denied area, for example, see Figure \ref{fig:commdeniedexamples}b.
For our experiments, we consider an environment of size $64 \times 64$ units with 36 agents in the swarm. Agents have a maximum communication range, $CR$, of 5 units.

\begin{algorithm}
\caption{Agent Behaviour}
\begin{algorithmic}[1]
\While{$t < t_{max}$}
    \State BroadcastHeartbeat()
    \State ReceiveHeartbeats()
    \State BroadcastAcknowledgements()
    \State $obs_{c} \gets$ ObserveCommunicationQuality($x,y$)
    \State ObservationsUpdate($obs_{c}$)
    \If{State $== E$}
        \State $obs_{f} \gets$ ObserveFeatures($x,y$)
        \State EstimateUpdate($obs_{f}$)
        \If{$t\%60 == 0$}
            \State State $=D$, $t=0$, $g=$ UpdateModulation()
        \EndIf
    \ElsIf{State $== D$}
        \State UpdateStoredBeliefs()
        \If{$t\%120g == 0$}
            \State State $=D$, $t=0$
        \EndIf
    \EndIf
    \State Planning(State)
    \State Movement(State)
    \State $t = t + 1$
\EndWhile
\end{algorithmic}
\label{alg:mastercode}
\end{algorithm}
\subsection{Agent Behaviour}

Algorithm \ref{alg:mastercode} outlines the main loop executed by agents at each timestep of our simulation up to $t_{max}$ timesteps. Agents spend their time in one of two states, these being the \textit{Dissemination} state (\textit{D}), and the \textit{Exploration} state (\textit{E}). 

In the \textit{D} state, agents broadcast their beliefs over the dominant feature in the environment to neighbouring agents within communication range. In this state, agents cannot observe the features in the environment. The time spent in the \textit{D} state is determined via a parameter, $g$, which is defined for each of the decision making strategies and may last for a maximum of 120 timesteps. Agents in the \textit{E} state, observe the features in the environment, but cannot disseminate their beliefs to neighbouring agents. The time spent in the \textit{E} state is fixed at 60 timesteps. When the agent has spent enough time in its current state, it switches to the other state (from \textit{D} to \textit{E} or vice versa) and resets its timestep counter, $t$.

At each timestep, agents estimate the communication link quality at their current location and use this to update their communication quality map, which is modelled using a Gaussian Process (GP). In order to estimate this value, agents make use of heartbeat messages that are broadcast at every timestep. These messages include the transmitting agent's beliefs over the features in the environment and their current state. While agents always transmit their beliefs, it will only be considered by a receiving agent if the sender's state is set as \textit{D}. When an agent receives a heartbeat message, they immediately send an acknowledgement message. Agents keep a record of the IDs of neighbours (assigned randomly at the start of the experiment), which they have received acknowledgements and heartbeats from for the current timestep. This allows an estimate of the probability that a message is successfully received from the agent's current location by using equation \ref{eq:commsestimate}.

\begin{equation}
    obs_c = \frac{|Acknowledgements|}{|Acknowledgements \cup Heartbeats|}
    \label{eq:commsestimate}
\end{equation}

This can be seen in lines 4 and 5 of Algorithm \ref{alg:mastercode}. Agents also perform path planning, based on their communication quality map, in both \textit{D} and \textit{E} states and use the result of this planning to move to a favourable location, depending on their state, seen in lines 18 and 19 of Algorithm \ref{alg:mastercode}.

\subsection{Decision Making Strategies}

We apply three collective decision making strategies to test our communication aware algorithm, which are taken from previous literature. These are \textit{Direct Comparison of Option Quality} (DC) \cite{valentini2016collective}, \textit{Direct Modulation of Majority-based Decisions} (DMMD) \cite{valentini2016collective, valentini2015efficient} and the feature decision making algorithm proposed by Ebert et al. which henceforth will be labelled as MFDM \cite{ebert2018multi}.

In DC, agents share their beliefs, $b_i \in \{0,1\}$, about which feature in the environment is dominant and also their current estimate of the ratio for the dominant feature, $\rho_i$, during the \textit{D} state. This information is broadcast to other agents that are within communication range of agent $i$. When $i$ receives a broadcast from another agent, $j$, containing $b_j$ and $\rho_j$, agent $i$ stores this information in a vector, $B$, where it is held until agent $i$ enters the \textit{E} state. In DC, the parameter $g$ shown in line 15 of Algorithm \ref{alg:mastercode} is equal to 1, meaning the \textit{D} state lasts for a fixed period of 120 timesteps in our simulation. When leaving the \textit{D} state, agents compare their current estimate of the dominant feature with an estimate, $\rho_j$, selected at random from $B$, shared by a neighbouring agent of the swarm with belief $b_j \in {0,1}$. If the estimate, $\rho_j$, is greater than the agent's own estimate $\rho_i$, then the agent alters its belief to $b_j$. This belief-update process is outlined in Algorithm \ref{alg:dcupdate}. During the \textit{E} state, which lasts for a fixed time period of 60 timesteps in our simulation, agents cannot broadcast their own beliefs or estimates of which feature is dominant in the environment, but are able to receive them from other agents and are limited to moving and sensing in the environment, while building their current estimate, ready for the next \textit{D} period.
    
\begin{algorithm}
\caption{DC Belief Update}
\begin{algorithmic}[1]
\Require $\rho_i$ Agent's current estimate, $B$ Beliefs and estimates of dominant feature received from neighbouring agents
\State $\rho_j$, $b_j \gets $ Random $ \rho, b \in B$
\If{$\rho_j > \rho_i$}
    \State $b_i \gets b_j$
\EndIf
\end{algorithmic}
\label{alg:dcupdate}
\end{algorithm}
    
In DMMD, agents are limited to sharing their beliefs during the \textit{D} state. Agents maintain an estimate of the ratio for their current belief which is used to modulate the time spent in the \textit{D} state. This is done by setting $g = \rho_i$. Thus, a higher estimate results in a longer \textit{D} state and a greater chance of influencing the belief of other members of the swarm. At the end of the \textit{D} state, agents sum the number of each belief, $b_j$, received during that \textit{D} state, including their own, and alter their belief to reflect the most frequent belief. 
    
\begin{algorithm}
\caption{DMMD Belief Update}
\begin{algorithmic}[1]
\Require $b_i$ Agent's current belief, $B$ Beliefs of all agents encountered including $b_i$
\State $n_1 \gets$ SUM$(b_j)$ if $b_j = 1 \mbox{ } \forall b_j \in B$  
\State $n_0 \gets$ SUM$(b_j)$ if $b_j = 0  \mbox{ } \forall b_j \in B$  
\If{$n_0 > n_1$}
    \State $b_i \gets 0$
\Else
    \State $b_i \gets 1$
\EndIf
\end{algorithmic}
\label{alg:dmmdupdate}
\end{algorithm}
    
Agents then enter the \textit{E} state for a fixed time period of 60 timesteps. This belief update is detailed in Algorithm \ref{alg:dmmdupdate}. MFDM also modulates the time spent in the \textit{D} period. In this strategy, $g$ is assigned as the maximum of an agent's estimates for both possible beliefs, $g = \textrm{max}\{\rho_1, \rho_0\}$ where $\rho_1$ and $\rho_0$ correspond to the estimates of the ratio of feature 1 and feature 0 respectively. This is labelled as the confidence of the agent. Agents with estimates close to $0$ or $1$ will have a high confidence that their belief is correct, while agents with estimates close to $0.5$ will have a lower confidence and as such will spend less time disseminating. During the \textit{D} state, agents share their estimates with neighbouring agents, and at the end of this state, agents take the most frequent estimate that they have received in that period as their belief, similar to DMMD, detailed in Algorithm~\ref{alg:dmmdupdate}. Agents also maintain a concentration value that is updated when an agent receives a belief from another agent that it has not been in contact with for 180 timesteps. This concentration term is defined as $C^* = 0.9C + 0.1b$ where $C$ is the current concentration and $b$ is the received belief. This acts as a long term memory of the agent's perceived belief of the swarm over time and makes this approach more robust to changes in the environment. When the concentration remains at $0.1$ or $0.9$ for 30 timesteps, in our simulation, the agent locks in its decision which is equal to the agent's current belief.

\subsection{Communication Mapping Algorithm}

We adopt a generic framework for considering the quality of service (QoS) of the communication links between agents. Since QoS metrics are application dependent and may be affected by environmental factors or the type of data being shared, we consider an approach that is independent of any specific application requirements. Our approach also allows for the model to take into account changes that may be observed, differing from the theoretical values. 

In our problem, agents have no \textit{a priori} information relating to the location or severity of the communication limitations in the environment and they need to build a decentralised representation of the communication environment. To do this, we propose a Gaussian Process (GP) model \cite{rasmussen}.

A GP is specified by its mean function, $\mu(x)$, and covariance function, $k(x, x')$ of a real process, $f(x)$, given by
\begin{equation}
    \mu (x) = \mathbb{E} [f(x)]
\end{equation}
\begin{equation}
    k(x, x') = \mathbb{E}[(f(x) - \mu(x))(f(x') - \mu(x'))]
\end{equation}
\begin{equation}
    f(x) \sim \mathcal{GP}(\mu(x), k(x, x')).
\end{equation}

The aim of a GP is to estimate the function, $f(x)$. Formally, let $P_i = [p_{1}^i, ... , p_{N}^i]$ be the vector of $N_i$ points in the environment with corresponding communication quality observations, $Q_i = [q_{c1}^i, ... , q_{cN}^i]$, made by agent $i$. Then given that $q = f_i(p)$, for a set of test locations, $P_{i}^* \subseteq P$, we can model the joint distribution of the training outputs, $f_i$ and the test outputs, $f_i^*$ as the following which assumes a mean of 0 and noise free observations:

\begin{equation}
    \left[ {\mycom{\bm{f_i}}{\bm{f_i^*}}} \right] \sim \mathcal{N} \left( 0, \left[ {\mycom{K(P_i, P_i)}{K(P_i^*, P_i)}} {\mycom{K(P_i, P_i^*)} {K(P_i^*, P_i^*)}} \right] \right)
\end{equation}

The covariance function specifies the joint variability between pairs of cells in the environment $p$ and $p'$, and for our approach we use the Mat\'ern kernel as the covariance function, defined as the following where $\nu$ is a smoothness parameter and $\ell$ is the length scale:

\begin{equation}
    k(p, p') = {\frac {2^{1-\nu }}{\Gamma (\nu )}}{\Bigg (}{\sqrt {2\nu }}{\frac {||p - p'||}{\ell}}{\Bigg )}^{\nu }K_{\nu }{\Bigg (}{\sqrt {2\nu }}{\frac {||p - p'||}{\ell}}{\Bigg )}
\end{equation}

Each agent, $i$, maintains a GP model which is learned using estimations of the communication link quality, which are gathered from the environment, $\{P_i, Q_i, T_i\}$ where $T_i$ is the time when the estimation was made. In our implementation, 100 estimations are kept. This is in the interest of reducing the computational complexity of updating the GP model. 



When an estimate of the communication link quality is made by agent $i$, Algorithm \ref{alg:waos} is performed to determine whether or not the estimate should be stored in $P_s$, to be used for the generation of the communication quality map. Agent $i$ calculates the minimum distances from all stored estimations, $p_s \in P_s$ to any other one of its stored estimations, $p_{s'} \in P_s$ s.t. $p_{s'} \neq p_s$. These distances are stored in $D^i_{min}$. Agent $i$ then calculates the minimum distance between the current estimation, $p_e$ and each of its stored estimations, $p_s \in P_s$. This process is shown in lines 2-6 of Algorithm \ref{alg:waos}. If the minimum distance for the current estimation, $p_e$ is greater than the smallest minimum distances in $D^i_{min}$, then one of the observations with this minimum distance is randomly selected to be replaced by $p_s$. This can be seen in lines 7-10 of Algorithm \ref{alg:waos}.
\begin{algorithm}
\caption{Observation sampling}
\begin{algorithmic}[1]
\Require $P_e$ Communication estimate, $P_s$ Stored estimations, $D_{min}^i$ Minimum distances for agent $i$
\State $min\_distance \gets \infty$
\For{$p_s \in P_s$}
    \If{$||p_e - p_s|| < min\_distance$ s.t. $p_s \neq p_e$}
        \State $min\_distance \gets ||p_s - p_e||$
    \EndIf
\EndFor
\If{$min\_distance > \textrm{min}(D_{min}^i)$}
    \State $P_s[\textrm{index\_of}(\textrm{min}(D_{min}^i))] \gets p_e$
    \State Update $D_{min}^i$
\EndIf
\end{algorithmic}
\label{alg:waos}
\end{algorithm}
Following this algorithm, agents aim to maximise the coverage of the estimations that are stored in their GP model and as such build up a representation of the communication environment, given a small number of data points. If the agent's stored estimations set is not at capacity (100 estimates), then the agent adds its most recent estimation of the communication environment. This allows for the wide area map to be initiated at the beginning of the task.

 
\subsection{Path Planning Algorithm}

Using the GP model, each agent, $i$ is able to assign an estimated communication quality, $\hat{q}_{x,y}^i$, and uncertainty over the modelled communication quality environment, $\hat{\sigma}_{x,y}^i$ for each point, ($x,y$) in the environment.

During the \textit{D} state, agents consider a square portion of their communication quality map, $A$, with the agent positioned at the centre. $A$ has sides of length $l$, in our simulation this is set to $l=10$. Agents then split this region into $n$ equally sized areas, $a^i_n \in A$, each with lengths of size $\frac{l}{\sqrt{n}}$, in our simulation $n=9$. The potential rewards for moving to each area in the \textit{D} state, $R_D(a^i_n)$, is given by the sum of the estimated communication quality at each point in that area.

\begin{equation}
    R_D(a^i_n) = \sum_{x,y \in a^i_n} \hat{q}_{x,y}^i
\end{equation}

A similar approach is used in the \textit{E} state, however instead of using a portion of the estimated communication quality map, agents use a portion of the uncertainty map. Hence, the potential rewards for moving to each area in the \textit{E} state, $R_O(a^i_n)$ is given by the following equation:

\begin{equation}
    R_O(a^i_n) = \sum_{x,y \in a^i_n} 1 - \hat{\sigma}_{x,y}^i
\end{equation}

Once an agent has determined which area holds the highest potential rewards, it moves in a straight line towards the centroid of that area. When the agent reaches the centroid the path planning algorithm is performed again and the agent moves from area to area based on the expected rewards. This is the planning algorithm without any coordination between agents, abbreviated to CA-G in our study.

\subsection{Agent Coordination}

We now introduce our swarm coordination algorithm which is implemented to reduce the possibility of agents moving to the same area, possible when agents only employ the CA-G algorithm. The intuition behind our coordination algorithm, CA-Co, is for  agents to maximise their possibility of successfully communicating during the \textit{D} state and to maximise coverage to reduce uncertainty over the environment in the \textit{E} state. Consider a single agent, $i$, in the \textit{D} state. When entering the planning state, agent $i$ uses the path planning algorithm to determine the potential rewards, $R_D(a^i_n)$ for moving to each of the $n$ areas, $a^i_n \in A$ within their considered portion of the communication quality map, $A$. Instead of moving to the centroid of the best area, as in the path planning algorithm, agent $i$ assigns a score to each of the areas, $s^i_n$, which reflects the contribution of each area, to the total reward of all considered areas, for agent $i$. This is given by
$s^i_n = \frac{R_D(a^i_n)}{\sum_{a^i_m \in A} R_D(a^i_m)}
$. Agents then choose a random area to move to using the calculated scores as weights. As such, areas with higher scores, and thus expected rewards, are more likely to be visited. 


\section{Results}

We compare our proposed communication aware algorithms, CA-Co and CA-G, with the DC, DMMD and MFDM decision making strategies. Results are compared with Random Baseline (RB), the original implementations of these decision making strategies, wherein agents move completely at random to explore and disseminate their beliefs. All experiment were replicated 20 times.
\begin{figure}
    \centering
    \includegraphics[trim={2cm 0cm 2cm 1cm},clip,width=0.47\textwidth]{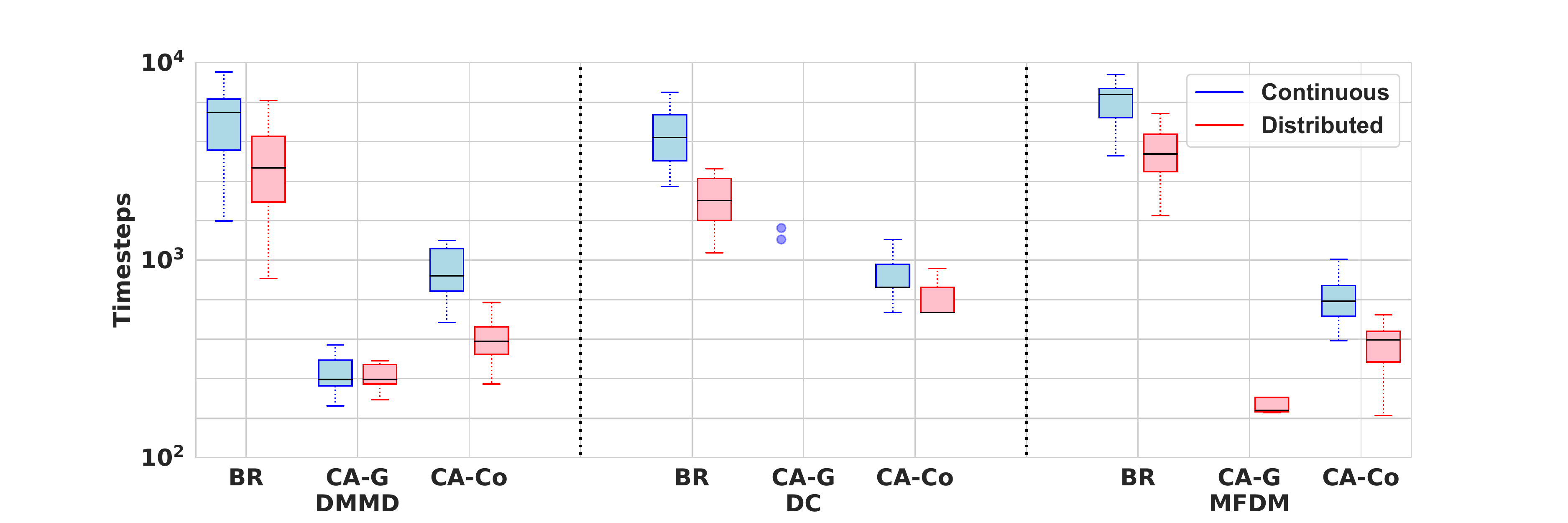}
    \caption{Time to convergence for each decision making strategy (DMMD, DC and MFDM), with $r_c=0.5$ and $r_f=0.65$ in continuous communication denied region (blue) and distributed communication denied region (red).}
    \label{fig:convergenceDC0.5}
\end{figure}

Figure \ref{fig:convergenceDC0.5} shows the time taken by the swarm to achieve consensus on the dominant feature in the environment, when agents of the swarm were denied communication in half the environment, $r_c=0.5$. Any replicates where the swarm failed to reach a consensus within 9000 steps, or reached an incorrect consensus are disregarded in the figures. The feature fill ratio for this experiment is $r_f = 0.65$. The results show that the CA approaches, regardless of the coordination (CA-G and CA-Co), greatly improve the speed at which the swarm converges to a consensus, i.e. each and every member of the swarm holds the same belief at a given timestep, with our proposed CA approaches achieving consensus over 300\% faster across all decision making strategies. The results suggest that the algorithm without coordination, CA-G has the best performance, particularly under the DMMD strategy, however, for the continuous communication constrained environment, the CA-G approach reached the incorrect consensus in 9 of the 20 replicates, compared to the CA-Co approach which was incorrect in 5 of the 20 replicates. All other runs for these approaches were successful. The BR approach failed to reach a consensus in 5 of the replicates. Another interesting observation here is that the CA-G approach failed to reach any consensus at all in all but 2 of the replicates that used the DC approach over both communication denied environments. This was caused by the agents forming clusters of agents over the course of the task that would very rarely mix. If all of the agents in one cluster disseminated beliefs that were incorrect, the swarm would not reach a consensus. This clustering issue was resolved by the CA-Co approach which successfully reached the correct consensus on all 20 replicates for both the continuous and distributed communication denied environments. 
\begin{figure}
    \centering
    \minipage{0.42\textwidth}
    \subfloat[]{\includegraphics[trim={1cm 0cm 1cm 1cm},clip,width=\linewidth]{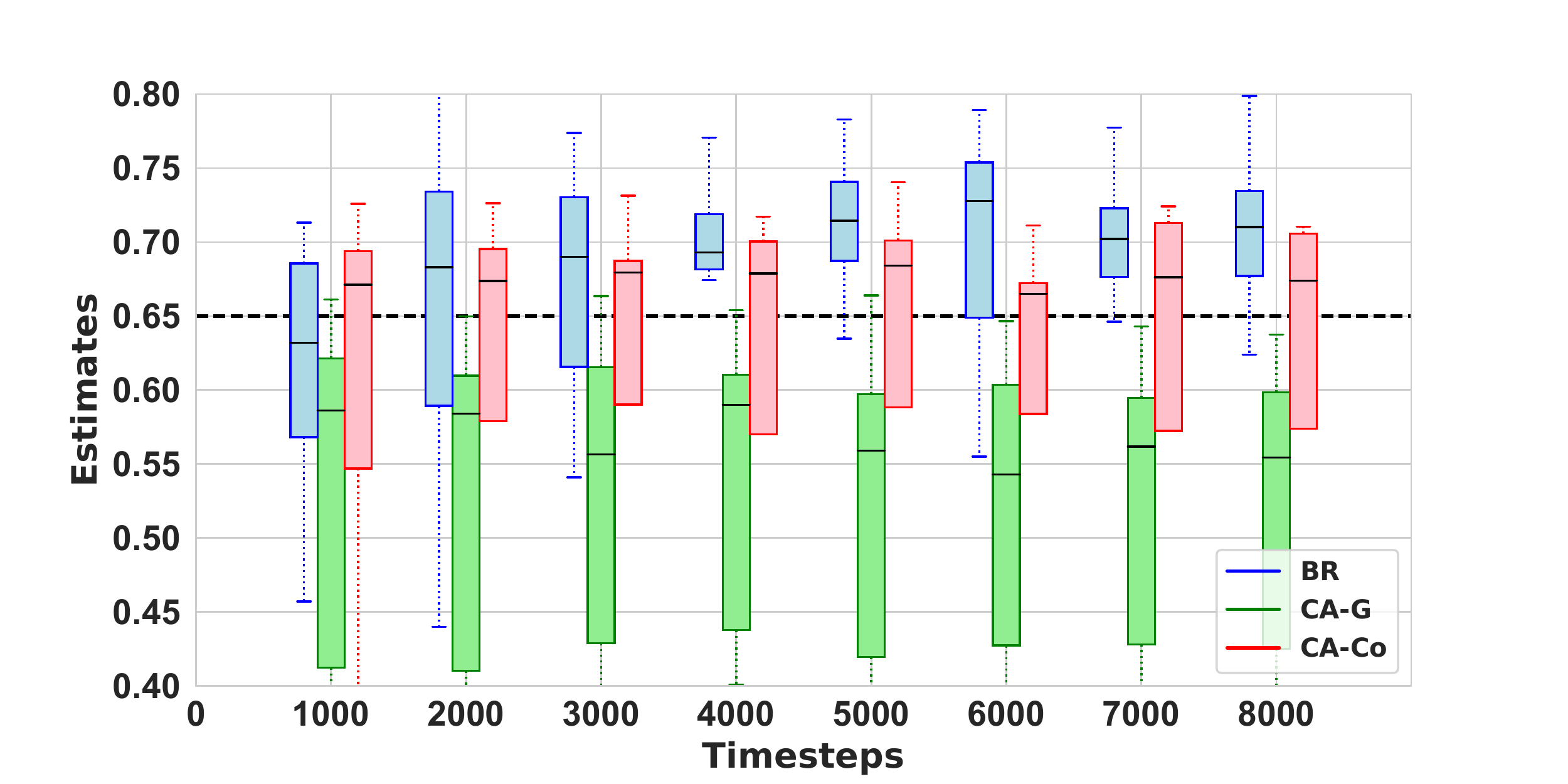}} \label{fig:contEstimates}
    \endminipage\hfill
    \minipage{0.42\textwidth}
    \subfloat[]{\includegraphics[trim={1cm 0cm 1cm 1cm},clip,width=\linewidth]{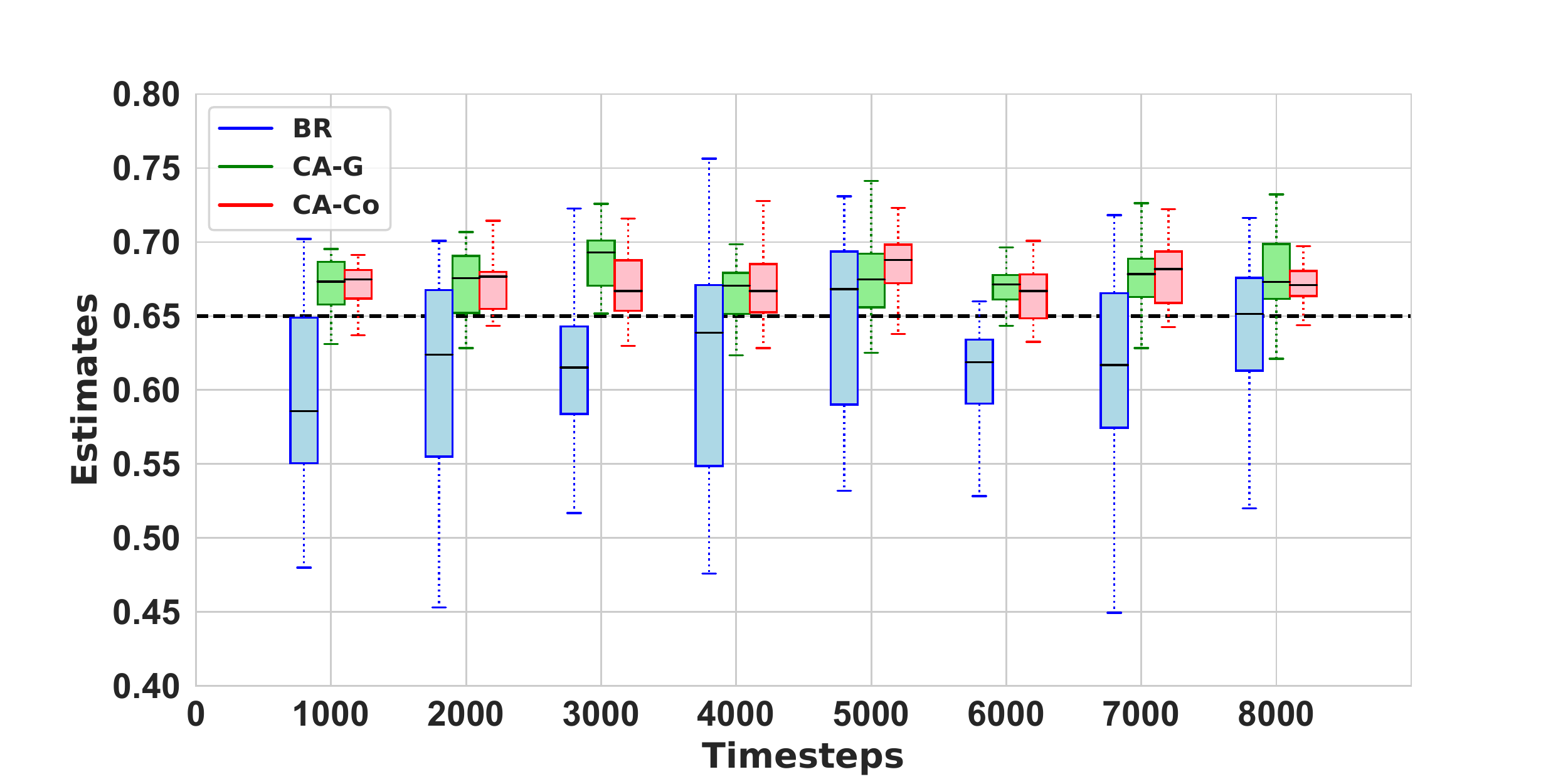}}
    \endminipage
    \caption{Estimates for DC decision making strategy in a (a) continuous environment and (b) distributed environment, with $r_{c}=0.5$ and $r_{f}=0.65$. The black dotted line indicates the ground truth.}
    \label{fig:estimatesDC0.5}
\end{figure}

While we have demonstrated that the speed of the decision is greatly increased by the communication aware approaches, another important factor is the accuracy of the decision that has been made. Figure \ref{fig:estimatesDC0.5} shows the estimates of the ratio of the dominant feature for the DC strategy with $r_c = 0.5$ and $r_f = 0.65$. It can be seen that the estimates for BR, CA-G and CA-Co are close to the ground truth, confirming that the swarm is on average estimating the ratio of the dominant feature in the environment, with a median error of no more than 13\%. Interestingly, while the CA-Co approach does not improve the time taken for the swarm to reach a consensus over the CA-G approach (Figure \ref{fig:convergenceDC0.5}), it does improve the median estimate in the distributed communication denied environment (Figure \ref{fig:estimatesDC0.5}b).  

\begin{figure}
    \centering
    \minipage{0.47\textwidth}
    \subfloat[]{\includegraphics[trim={2cm 0cm 2cm 1cm},clip,width=\linewidth]{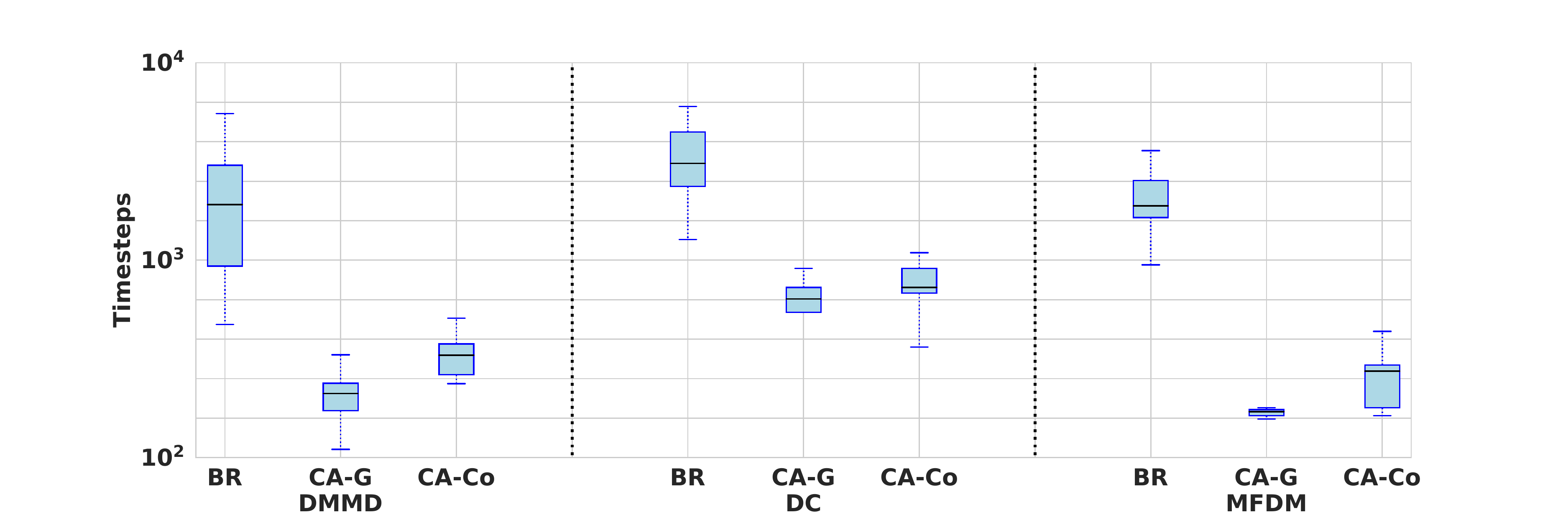}} \label{fig:noComm}
    \endminipage\hfill
    \minipage{0.47\textwidth}
    \subfloat[]{\includegraphics[trim={2cm 0cm 2cm 1cm},clip,width=\linewidth]{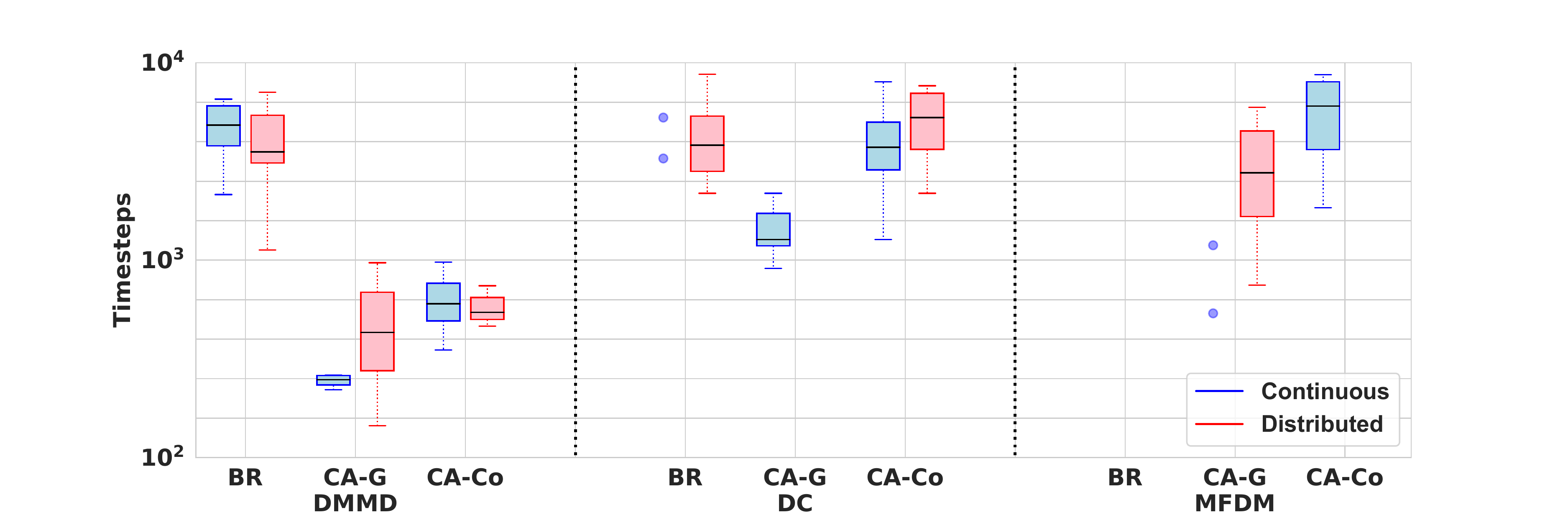}}
    \endminipage
    \caption{Time to convergence for each decision making strategy, (DC, DMMD and MFDM) with (a) no communication denied regions present and (b) $r_c = 0.5$ and $r_f = 0.53$ in continuous communication denied region (blue) and distributed communication denied region (red).}
    \label{fig:convergenceNoCommandHarder}
\end{figure}

We also tested the time taken for the swarm to reach a consensus for a feature ratio of $r_f = 0.65$ and communication environment with $r_c = 0$, meaning there are no communication denied areas in the environment and at any point in the environment, an agent is certain that any messages sent will be received by any agents in communication range. This mimics the task proposed by Ebert et al. \cite{ebert2018multi}. Figure \ref{fig:convergenceNoCommandHarder}a) gives the results for this experiment. They show that the fully coordinated approach continues to outperform the RB strategy. Not only is the median time to convergence considerably faster, the spread of the time taken is greatly reduced, making the CA-Co approach far more predictable and reliable for a collective decision making scenario. This is due to the agents exhibiting some level of aggregation when in the dissemination state with agents moving towards the same regions as some of their neighbours. This allows for faster mixing of agents' beliefs and thus reach a consensus more quickly. In the BR approach, this aggregation is not present and as such, the beliefs are spread around the swarm more slowly.

Finally, we increased the difficulty of the task by using environments with a ratio of the dominant feature, $r_f$, closer to 0.5, in this case 0.53. This presents a more challenging task since it is harder for individual members of the swarm to individually distinguish which feature is the most dominant at any point. Figure~\ref{fig:convergenceNoCommandHarder}b) shows that the CA-Co approach improves the median time taken to converge to a decision over the BR approach. The CA-Co approach also improves the reliability for the swarm to converge towards a consensus in this scenario. when using the MFDM strategy for decision making, where none of the BR replicates reached any concensus at all. In contrast, the CA-Co approach reached the correct consensus in 13 of the 20 replicates for the continuous communication denied environment, failing to reach a consensus in the other 7 replicates. The CA-G approach also reached the correct decision in 9 of the 20 replicates for the distributed communication environment, failing to reach any consensus at all in the other 11 replicates.

\section{Conclusion}

We present communication aware algorithms for a swarm to perform collective decision making in harsh communication limited environments. Our approach has been demonstrated to improve the ability for a swarm to reach a consensus on the feature of interest in the environment, not only in communication limited environments, but also in environments where the communication conditions are homogeneous across the environment. This improves on random movement which is a common method for determining how agents may move in collective decision making problems. Not only does our communication-aware approach reduce the time it takes the swarm to reach a consensus, but this is achieved while retaining the accuracy of the decision. We have not performed an extensive search for parameter selection and this is left for future work. We further plan to investigate the effect of dynamic communication-denied areas that change in number and size during the experiments.

\bibliographystyle{IEEEtran}
\bibliography{root.bib}

\begin{thebibliography}{10}
\providecommand{\url}[1]{#1}
\csname url@samestyle\endcsname
\providecommand{\newblock}{\relax}
\providecommand{\bibinfo}[2]{#2}
\providecommand{\BIBentrySTDinterwordspacing}{\spaceskip=0pt\relax}
\providecommand{\BIBentryALTinterwordstretchfactor}{4}
\providecommand{\BIBentryALTinterwordspacing}{\spaceskip=\fontdimen2\font plus
\BIBentryALTinterwordstretchfactor\fontdimen3\font minus
  \fontdimen4\font\relax}
\providecommand{\BIBforeignlanguage}[2]{{%
\expandafter\ifx\csname l@#1\endcsname\relax
\typeout{** WARNING: IEEEtran.bst: No hyphenation pattern has been}%
\typeout{** loaded for the language `#1'. Using the pattern for}%
\typeout{** the default language instead.}%
\else
\language=\csname l@#1\endcsname
\fi
#2}}
\providecommand{\BIBdecl}{\relax}
\BIBdecl

\bibitem{valentini2017best}
G.~Valentini, E.~Ferrante, and M.~Dorigo, ``The best-of-n problem in robot
  swarms: Formalization, state of the art, and novel perspectives,''
  \emph{Frontiers in Robotics and AI}, vol.~4, p.~9, 2017.

\bibitem{hamann2018swarm}
H.~Hamann, \emph{Swarm robotics: A formal approach}.\hskip 1em plus 0.5em minus
  0.4em\relax Springer, 2018.

\bibitem{roundtree2019transparency}
K.~A. Roundtree, M.~A. Goodrich, and J.~A. Adams, ``Transparency: Transitioning
  from human--machine systems to human-swarm systems,'' \emph{Journal of
  Cognitive Engineering and Decision Making}, vol.~13, no.~3, pp. 171--195,
  2019.

\bibitem{valentini2016collective}
G.~Valentini, D.~Brambilla, H.~Hamann, and M.~Dorigo, ``Collective perception
  of environmental features in a robot swarm,'' in \emph{International
  Conference on Swarm Intelligence}.\hskip 1em plus 0.5em minus 0.4em\relax
  Springer, 2016, pp. 65--76.

\bibitem{valentini2014self}
G.~Valentini, H.~Hamann, M.~Dorigo \emph{et~al.}, ``Self-organized collective
  decision making: the weighted voter model.'' in \emph{AAMAS}, 2014, pp.
  45--52.

\bibitem{valentini2016100kilobots}
G.~Valentini, E.~Ferrante, H.~Hamann, and M.~Dorigo, ``Collective decision with
  100 kilobots: Speed versus accuracy in binary discrimination problems,''
  \emph{Autonomous agents and multi-agent systems}, vol.~30, no.~3, pp.
  553--580, 2016.

\bibitem{ebert2018multi}
J.~T. Ebert, M.~Gauci, and R.~Nagpal, ``Multi-feature collective decision
  making in robot swarms,'' in \emph{Proceedings of the 17th International
  Conference on Autonomous Agents and MultiAgent Systems}, 2018, pp.
  1711--1719.

\bibitem{kernbach2013adaptive}
S.~Kernbach, D.~H{\"a}be, O.~Kernbach, R.~Thenius, G.~Radspieler, T.~Kimura,
  and T.~Schmickl, ``Adaptive collective decision-making in limited robot
  swarms without communication,'' \emph{The International Journal of Robotics
  Research}, vol.~32, no.~1, pp. 35--55, 2013.

\bibitem{schmickl2011beeclust}
T.~Schmickl and H.~Hamann, ``Beeclust: A swarm algorithm derived from
  honeybees,'' \emph{Bio-inspired computing and communication networks}, pp.
  95--137, 2011.

\bibitem{mathews2010establishing}
N.~Mathews, A.~L. Christensen, E.~Ferrante, R.~O'Grady, and M.~Dorigo,
  ``Establishing spatially targeted communication in a heterogeneous robot
  swarm.'' in \emph{AAMAS}, 2010, pp. 939--946.

\bibitem{mathews2012spatially}
N.~Mathews, A.~L. Christensen, R.~O'Grady, and M.~Dorigo, ``Spatially targeted
  communication and self-assembly,'' in \emph{2012 IEEE/RSJ International
  Conference on Intelligent Robots and Systems}.\hskip 1em plus 0.5em minus
  0.4em\relax IEEE, 2012, pp. 2678--2679.

\bibitem{talamali2021less}
M.~S. Talamali, A.~Saha, J.~A. Marshall, and A.~Reina, ``When less is more:
  Robot swarms adapt better to changes with constrained communication,''
  \emph{Science Robotics}, vol.~6, no.~56, p. eabf1416, 2021.

\bibitem{coquet2021control}
C.~Coquet, A.~Arnold, and P.-J. Bouvet, ``Control of a robotic swarm formation
  to track a dynamic target with communication constraints: Analysis and
  simulation,'' \emph{Applied Sciences}, vol.~11, no.~7, p. 3179, 2021.

\bibitem{winfield2006safety}
A.~F. Winfield and J.~Nembrini, ``Safety in numbers: fault-tolerance in robot
  swarms,'' \emph{International Journal of Modelling, Identification and
  Control}, vol.~1, no.~1, pp. 30--37, 2006.

\bibitem{valentini2015efficient}
G.~Valentini, H.~Hamann, and M.~Dorigo, ``Efficient decision-making in a
  self-organizing robot swarm: On the speed versus accuracy trade-off,'' in
  \emph{Proceedings of the 2015 International Conference on Autonomous Agents
  and Multiagent Systems}, 2015, pp. 1305--1314.

\bibitem{rasmussen}
C.~Rasmussen and C.~Williams, \emph{Gaussian Processes for Machine
  Learning}.\hskip 1em plus 0.5em minus 0.4em\relax MIT Press, 2006.

\end{thebibliography}

\end{document}